\documentclass[runningheads,a4paper]{llncs}

\usepackage{amssymb}
\usepackage{graphicx}
\usepackage{amsmath}
\usepackage{tabularx}
\usepackage{multirow}

\usepackage{url}
\urldef{\mailsa}\path|{shwang, sgpark}@lunit.io|    
\newcommand{\keywords}[1]{\par\addvspace\baselineskip
\noindent\keywordname\enspace\ignorespaces#1}
\newcolumntype{Y}{>{\centering\arraybackslash}X}

\begin{document}

\mainmatter  % start of an individual contribution

%\title{Accurate Lung Segmentation with Network-Wise Training of Deep Convolutional Neural Networks}
\title{Accurate Lung Segmentation via Network-Wise Training of Convolutional Networks}

 \titlerunning{Lung Segmentation via Network-Wise Training}

\author{Sangheum Hwang\and Sunggyun Park}
\authorrunning{S. Hwang and S. Park}

\institute{Lunit Inc., Seoul, Korea\\
\mailsa}


\toctitle{Lecture Notes in Computer Science}
\tocauthor{Authors' Instructions}
\maketitle

\begin{abstract}

We introduce an accurate lung segmentation model for chest radiographs based on deep convolutional neural networks. Our model is based on atrous convolutional layers to increase the field-of-view of filters efficiently. To improve segmentation performances further, we also propose a multi-stage training strategy, \textit{network-wise training}, which the current stage network is fed with both input images and the outputs from pre-stage network. It is shown that this strategy has an ability to reduce falsely predicted labels and produce smooth boundaries of lung fields. We evaluate the proposed model on a common benchmark dataset, JSRT, and achieve the state-of-the-art segmentation performances with much fewer model parameters.

\end{abstract}

\section{Introduction}

%Segmentation of lung fields in chest radiographs is an important step for a computer-aided diagnosis (CAD) of lung health. 
Accurate lung boundaries provide valuable image-based information such as total lung volume or shape irregularities, but it also has an important role as a prerequisite step for developing computer-aided diagnosis (CAD) system. However, an automated segmentation of lung fields is a challenging problem due to high variations in shape and size among different chest radiographs. 

For automatic detection of lung fields, a lot of methods have been proposed over the past decade~\cite{Candemir2014TMI,Ginneken2006MIA,Ngo2015ICIP,Novikov2017arxiv}. The early segmentation methods can be partitioned into \textit{rule-based}, \textit{pixel classification-based}, \textit{deformable model-based}, and \textit{hybrid} methods~\cite{Ginneken2006MIA}. Recently, deep neural network-based approaches~\cite{Ngo2015ICIP,Novikov2017arxiv}\footnote{In~\cite{Ngo2015ICIP}, the authors propose a hybrid model combined distance regularized level sets with a deep learning model for lung segmentation. This model shows high overlap scores but it requires good initial guesses. Therefore, we exclude this model from our comparison.} have been proposed due to the success of deep learning in various computer vision tasks including object classification~\cite{Krizhevsky2012Classification}, localization~\cite{fasterrcnn2015nips}, and segmentation~\cite{deeplab2016arxiv,Noh2015iccv}.

For semantic segmentation, the encoder-decoder architecture is commonly used~\cite{Noh2015iccv}. In this architecture, encoder is a typical convolutional neural network (CNN), while decoder consists of transposed convolutions and upsampling operations. The role of decoder is to restore the abstracted feature map by learning how to densify the sparse activations. The final output of decoder is a probability map with the same size as that of the ground-truth masks, and pixel-wise cross entropy loss is employed for training. 
Such encoder-decoder architecture has also been shown its promising performances in various medical imaging problems~\cite{Novikov2017arxiv,unet_seg_1}. For example, U-Net~\cite{unet_seg_1}, a variant of the encoder-decoder architecture, shows the impressive results on segmentation of neuronal structures in electron microscopic stacks. For the task of lung segmentation, the authors of \cite{Novikov2017arxiv} present U-Net-based CNN architecture for automated segmentation of anatomical organs (e.g., lung, cavicles and heart) in chest radiographs. They also propose a modified loss function to deal with the multi-class segmentation problem. 

Another succesful approach for semantic segmentation is to employ atrous convolutional layers by replacing some convolutional layers~\cite{deeplab2016arxiv}. It is known that atrous convolution effectively enlarges the global receptive field of CNN~\cite{Luo2016nips}, and therefore larger context information can be efficiently utilized for prediction of pixel-wise labels. 

In this paper, we introduce an accurate lung segmentation model for chest radiographs based on deep CNN with atrous convolutions. The proposed model is designed to have a deep-and-thin architecture, which has much fewer parameters compared to other CNN-based lung segmentation models. To improve further, we propose a multi-stage training strategy, \textit{network-wise training}, which the current stage network is fed with both input images and the outputs of pre-stage network. It is shown that this strategy has an ability to reduce falsely predicted labels (i.e. false positives and false negatives) and produce smooth boundaries of segmented lung fields. 

We evaluate the proposed method on a common benchmark dataset, the Japanese Society of Radiological Technology (JSRT)~\cite{jsrt2000arj}, and achieve the state-of-the-art results under four popular segmentation metrics: the Jaccard similarity coefficient, Dice's coefficient, average contour distance, and average surface distance. To investigate generalization capability of our method, we test on another dataset, the Montgomery County (MC)~\cite{Jaeger2014TB}. It is observed that performances on this dataset are comparable in terms of mean values, but have high variances since there is some degree of a shift between training (JSRT) and test (MC) distributions.

\begin{figure}
\begin{center}
\includegraphics[width=\textwidth]{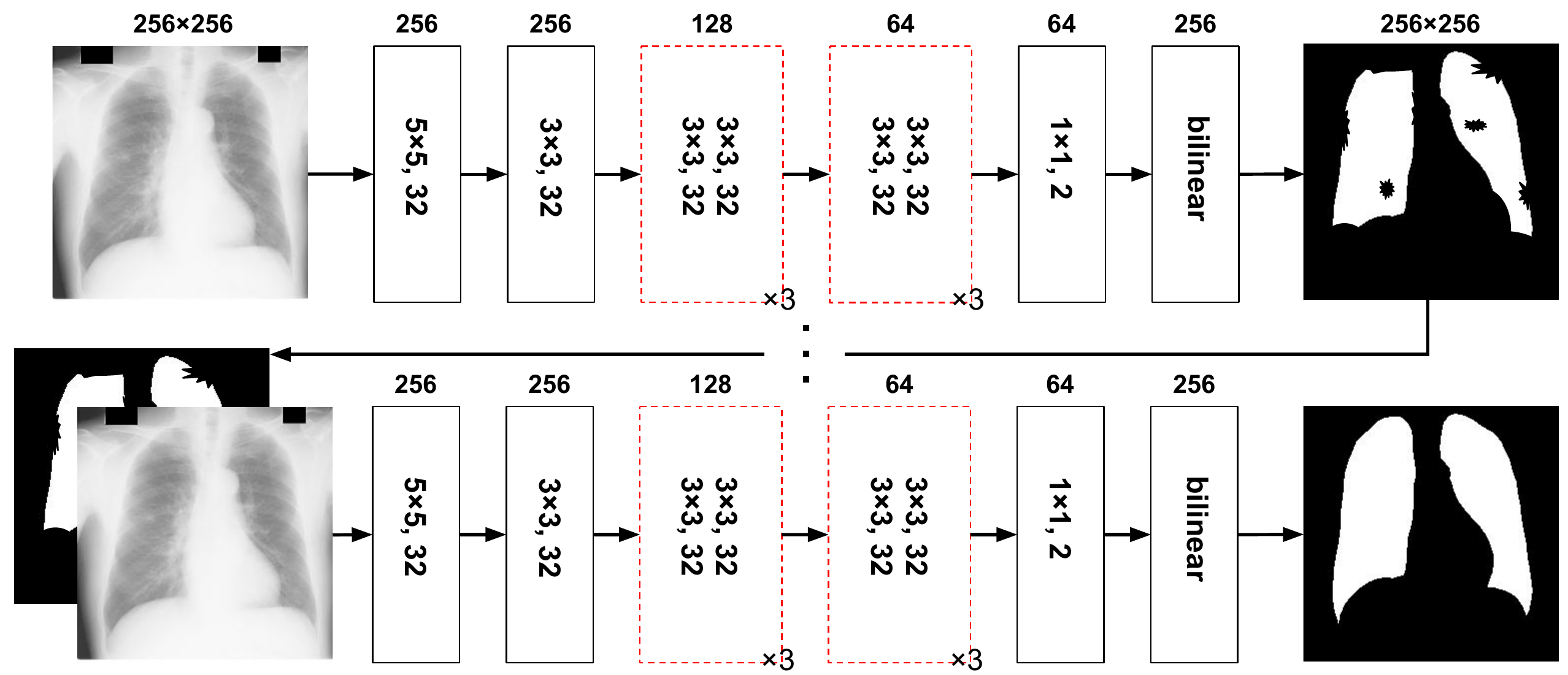}
\end{center}
%\vspace{-2mm}
   \caption{Simplified framework of the proposed method. In network, the boxes in black represent convolutions and bilinear interpolation, and the dotted boxes in red denote residual blocks consisting of consecutive two convolutional layers. The number in each box means the size and the number of filters, respectively, and that above each box represents the resolution of feature maps. Our model is trained in a network-wise manner by feeding the outputs from pre-stage network.}
\label{fig:overall_arch}
\end{figure}

\section{Methods}

\subsection{Lung Segmentation with Atrous Convolutions}
We present a deep-and-thin CNN architecture based on residual learning~\cite{resnet2016cvpr} which has skip connections to prevent the gradient vanishing problem. Dense prediction problems should consider large context to predict class labels of pixels. Simple way for larger context is increasing the global receptive fields of network by stacking more convolution layers or using downsampling operations (e.g., pooling or strided convolution)~\cite{Luo2016nips}.

Recently, it is known that atrous convolution is useful to enlarge the field-of-view (i.e. receptive fields) of filters. This enlargement is particularly effective for segmentation task since it should consider the context around the location where we want to predict class labels~\cite{deeplab2016arxiv}. Atrous convolution contains `holes' between weights of filters so that it involves larger fields to compute activations. Given a filter $\mathbf{k}=[k(m,n)]$ for $m,n=1,...,2k+1$ and the input $\mathbf{x}=[x(i,j)]$ at location $(i,j)$, atrous convolution with rate $r$ computes the output $\mathbf{y}=[y(i,j)]$ as follows:

\begin{equation}
\label{eq:atrous}
y(i,j) = \sum_{m=-k}^{k} \sum_{n=-k}^{k} {x(i+rm,j+rn)k(m,n)}
\end{equation}

\noindent Note that if $r=1$, Eq.~\ref{eq:atrous} stands for standard convolution operation. Therefore, the global receptive field of network can be controlled via rate $r$ while maintaining the number of weights.

Fig.~\ref{fig:overall_arch} shows an architecture of our network for lung segmentation task. It consists of 3 convolutional layers and 6 residual blocks, i.e. 15 convolutional layers. We employ atrous convolutions with $r=3$ for the end of two residual blocks. Batch normalization layer is followed by every convolutional layer. The global stride of our network is 4, i.e. 2 convolutional layers at the beginning of particular residual blocks (the first layer in each red block in Fig.~\ref{fig:overall_arch}) operate convolutions with stride 2 (i.e. 2-strided convolution). To calculate pixel-wise cross entropy loss with the groud-truth mask, we upsample network outputs by bilinear interpolation. 

The advantage of the proposed deep-and-thin architecture is that it has much fewer model parameters compared to other CNN-based lung segmentation models. For examples, our model has 120,672 weights (26 times fewer parameters) while the encoder-decoder network like U-Net has 3,140,771 weights~\cite{Novikov2017arxiv}.

\subsection{Network-wise Training of CNN}

Generally, CNN with atrous convolutions and bilinear interpolation has some limitations. First, it may produce small false positive or false negative areas. This is mainly caused due to pixel-wise cross entropy loss dealing with every pixel independently. Second, it outputs blurry object boundaries, which is inevitable if we use a bilinear interpolation to upsample the downsampled feature maps. To overcome these issues, postprocessing via conditional random fields~\cite{fccrf2011nips} is widely used to smooth such noisy segmentation maps~\cite{deeplab2016arxiv}.

We propose another strategy, network-wise training, to refine segmentation results. It is designed as a repeated training pipeline which has an output of pre-stage model as an input (see Fig.~\ref{fig:overall_arch}). At the first stage (namely stage 1), a network is trained using only input chest radiographs. After training it, both input chest radiographs and network outputs from trained model at stage 1 are fed into the second network. Specifically, input chest radiographs and the corresponding output from pre-stage network are concatenated across the channel dimension. From relatively coarse segmentation outputs, a network can more focus on the details to learn accurate boundaries of lung fields. This procedure is iterated until validation performance is saturated. Note that this strategy can be considered as iterative cascading, an extended version of the cascaded network~\cite{Havaei2017MIA}.
%From experiments, it is observed that it needs small number of iterations to converge.

\section{Computational Experiments}

We use a common benchmark dataset, the Japanese Society of Radiological Technology (JSRT) dataset~\cite{jsrt2000arj}, to evaluate lung segmentation performance of our model. JSRT database contains 247 the posterior-anterior (PA) chest radiographs, 154 have lung nodules and 93 have no nodules. The ground-truth lung masks can be obtained in the Segmentation in Chest Radiographs (SCR) database~\cite{Ginneken2006MIA}. 

Following previous practices in literatures, JSRT dataset is split in two folds: one contains 124 odd numbered and the other contain 123 even numbered chest radiographs. Then, one fold is used for training\footnote{After the search of hyperparameters with randomly selected 30\% training data, the network is re-trained with the entire training data.} and the other fold used for testing, and vice versa. Final performances are computed by averaging results from both cases. Also, all training images are resized to 256$\times$256 as in the literatures. The network is trained via stochastic gradient descent with momentum 0.9. For learning rate scheduling, we set initial learning rate to 0.1 and it is decreased to 0.01 after training 70 epochs.

We use Montgomery County (MC) dataset~\cite{Jaeger2014TB} as another testset to investigate generalization capability of our model. MC dataset contains PA chest radiographs collected from National Library of Medicine, National Institutes of Health, Bethesda, MD, USA. It consists of 80 normal and 58 abnormal cases with manifestations of tuberculosis. It is interesting to see segmentation performances on this dataset since it has different characteristics compared to training set (JSRT): image acquisition equipment, abnormal diseases, nationality of patients, etc.

\subsection{Performance Metrics}
We use four commonly used metrics in the literatures: the Jaccard similarity coefficient(JSC), Dice's coefficient (DC), average contour distance (ACD), average surface distance (ASD)\footnote{Average surface distance is also known as symmetric mean absolute surface distance~\cite{Novikov2017arxiv}.}. JSC and DC are similar in that they only consider the number of true positives, false positives and false negatives. Therefore, they are metrics ignoring predicted locations. On the other hand, ACD and ASD are distance-based metrics. They penalize if the minimum distance of a particular pixel predicted as lung boundaries to the ground-truth boundaries is large. Therefore, performance from these metrics may vary even if JSC and DC are almost the same.

Let $s_i$, $i=1,...,n_S$, and $g_j$, $i=1,...,n_G$, be the pixels on the segmented boundary $S$ and the ground-truth boundary $G$. The minimum distance of $s_i$ on $S$ to $G$ is defined as $d(s_i,G)=\min_j \| g_j - s_i \|$.  Then, ACD and ASD are computed as follows:

\begin{equation}
\begin{aligned}
\label{eq:acd_asd}
\text{ACD(S,G)}&=\frac{1}{2}\left(\frac{\sum_i d(s_i,G)}{n_S} + \frac{\sum_j d(g_i,S)}{n_G} \right) \\
\text{ASD(S,G)}&=\frac{1}{n_S+n_G}\left(\textstyle  \sum_i d(s_i,G) + \sum_j d(g_i,S) \right).
\end{aligned}
\end{equation}

\begin{figure}[t]
	\begin{center}
		\includegraphics[width=\textwidth]{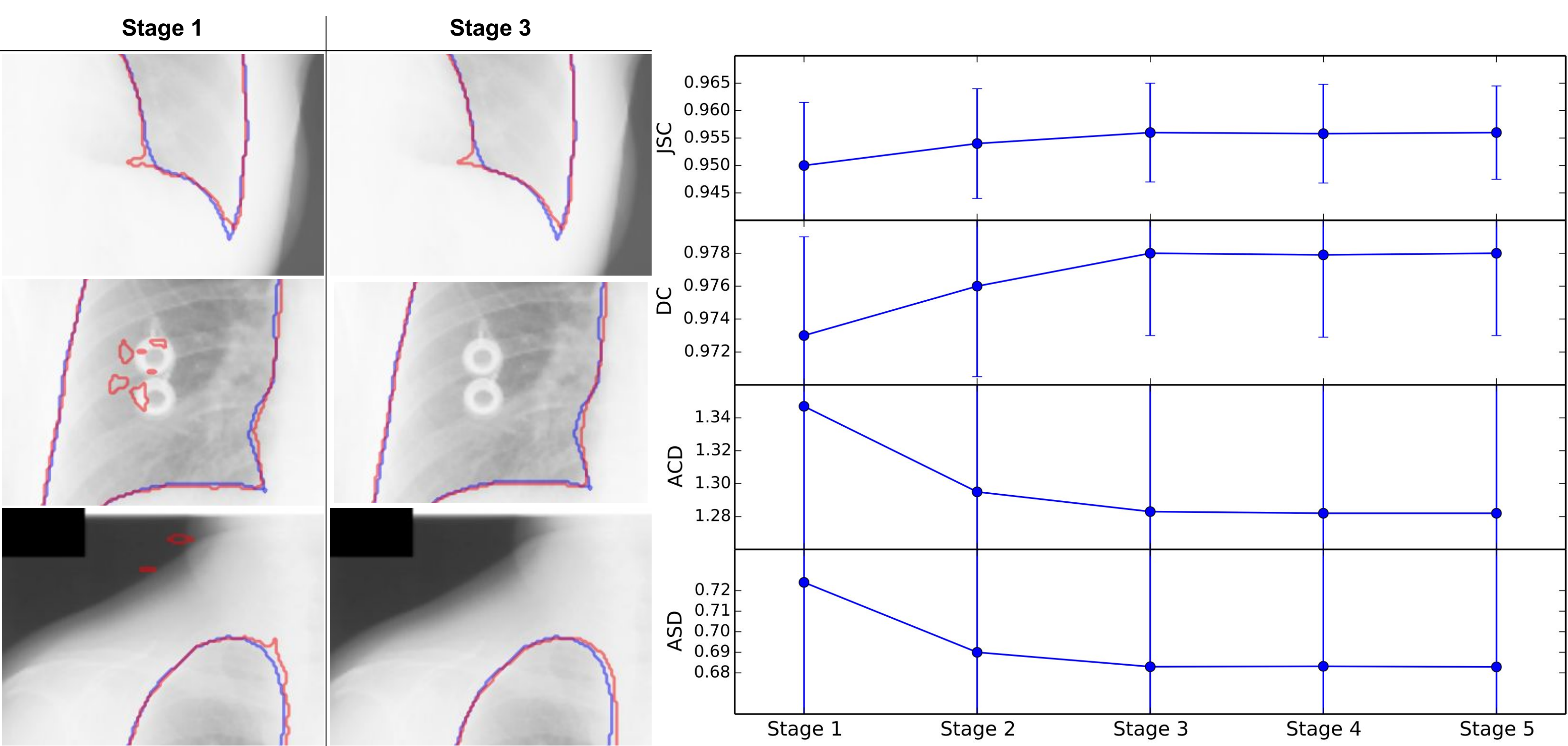}
	\end{center}
	%\vspace{-2mm}
	\caption{The effect of network-wise training at each stage. 1) Left figure shows some regions improved through network-wise training. The ground-truth and prediction is depicted in blue and red line, respectively. 2) Right plot shows performances at each stage. It is shown that they are saturated at stage 3.}
	\label{fig:stage}
\end{figure}

\begin{table}[t]
	\caption{Mean and standard deviations of segmentation performances for JSRT and MC datasets. The best mean performances for each dataset are given in bold.}
	\centering
	\label{table:comparison}
	\def\arraystretch{1.5}
	\scriptsize
	\begin{tabular}{c|l|cccc}
		\hline\hline
		%		\multicolumn{1}{c|}{}   & JSC         & DC          & ACD         & ASD  \\ \hline
		Dataset & Methods   & JSC         & DC          & ACD (\textit{mm})         & ASD (\textit{mm}) \\ \hline
		\multirow{9}{*}{JSRT} & Human observer~\cite{Ginneken2006MIA}         & 0.946$\pm$0.018 & -                       & 1.64$\pm$0.69       & -    \\
		& Hybrid voting~\cite{Ginneken2006MIA}              & 0.949$\pm$0.020 & -                      & 1.62$\pm$0.66       & -    \\
		& Candemir~\cite{Candemir2014TMI}                     & 0.954$\pm$0.015 & 0.967$\pm$0.008 & 1.321$\pm$0.316   & -    \\
		%& U-Net~\cite{Novikov2017arxiv}                           & 0.939               & 0.968               & -                       & 0.87 \\
		& InvertedNet~\cite{Novikov2017arxiv}                  & 0.950                 & 0.973              & -                       & 0.69 \\ \cline{2-6}
		& Proposed (Stage 1)                                         & 0.950$\pm$0.023     & 0.974$\pm$0.012    & 1.347$\pm$0.919  &   0.724$\pm$0.163   \\
		& Proposed (Stage 2)                                         & 0.954$\pm$0.020     & 0.976$\pm$0.011    & 1.295$\pm$0.846  &   0.690$\pm$0.151   \\
		& Proposed (Stage 3)                                         & 0.956$\pm$0.018     & 0.977$\pm$0.010    & 1.283$\pm$0.814  &   0.683$\pm$0.145   \\
		& Proposed w/ aug (Stage 3)                              &  \textbf{0.961}$\pm$0.015    & \textbf{0.980}$\pm$0.008   & \textbf{1.237}$\pm$0.702   &  \textbf{0.675}$\pm$0.122   \\ \hline
		\multirow{2}{*}{MC} & Candemir~\cite{Candemir2014TMI}         & \textbf{0.941}$\pm$0.034 & 0.960$\pm$0.018  & \textbf{1.599}$\pm$0.742       & -    \\ \cline{2-6}
		& Proposed w/ aug (Stage 3)                      & 0.931$\pm$0.049 & \textbf{0.964}$\pm$0.028 & 2.186$\pm$1.795   & 0.915$\pm$0.258    \\		
		\hline\hline
	\end{tabular}
\end{table}

\begin{figure}[t]
	\begin{center}
		\includegraphics[width=\textwidth]{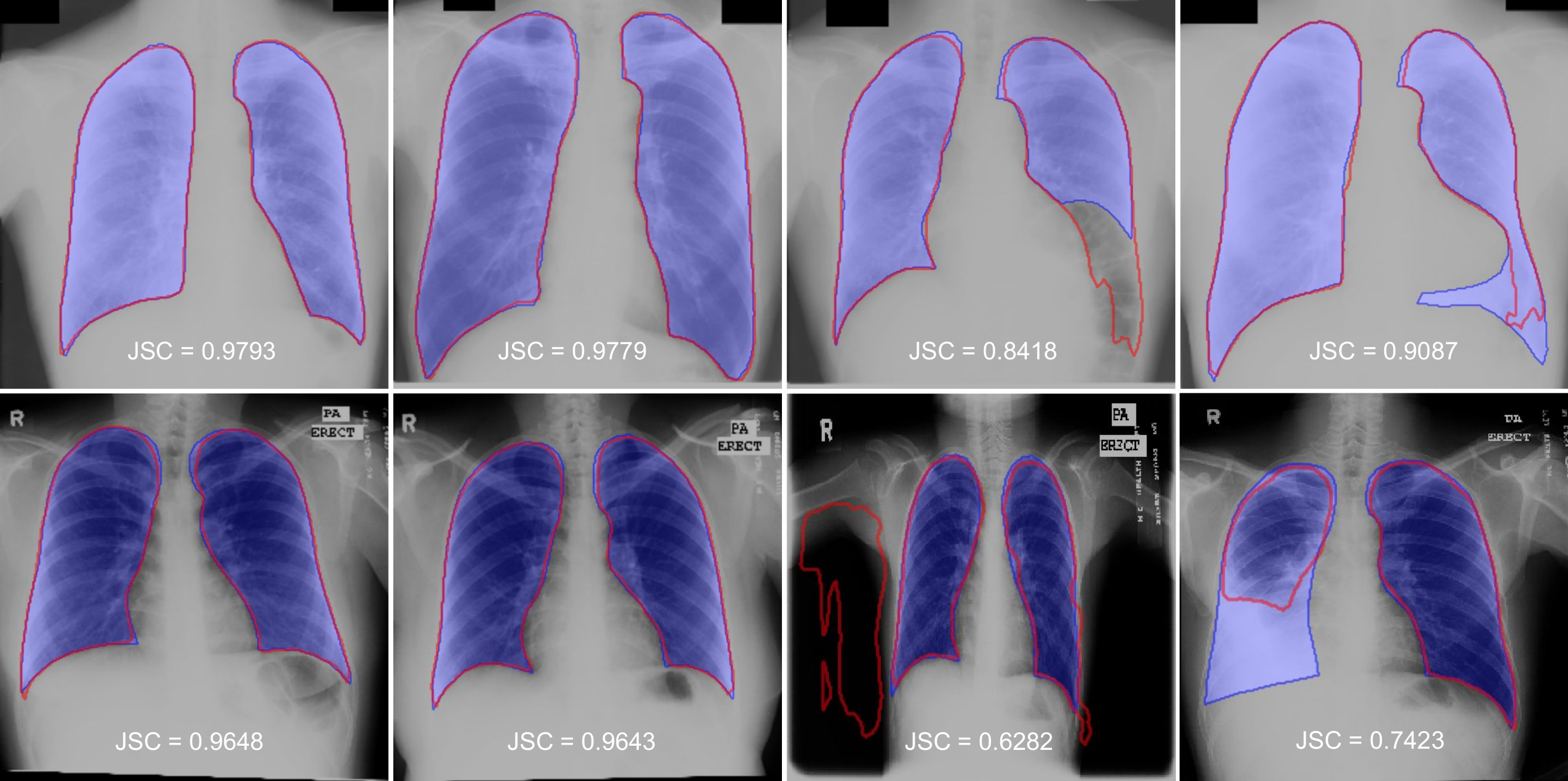}
	\end{center}
	%\vspace{-2mm}
	\caption{Segmentation examples for JSRT and MC datasets. The ground-truth masks are depicted in blue and segmentation boundaries from our model are in red. The first and second row show the results from JSRT and MC, respectively.	Left two columns and right two columns present the best and worst two examples in terms of JSC.}
	\label{fig:examples}
\end{figure}

\subsection{Quantatitive and Qualititive Results}
Table~\ref{table:comparison} summarizes segmentation performances of our model compared to previous methods\footnote{Note that JSC and DC numbers in Candemir~\cite{Candemir2014TMI} are incorrect since DC should be 2JSC/(1+JSC).}. First, we evaluate the models at stage 1 and 3, which are trained without any preprocessing method such as histogram equalization and data augmentation techniques to exclude other potential factors that may affect performances. These results show that segmentation performances are continously improved through a network-wise training, and those from stage 3 model outperforms other methods.  

The left side in Fig.~\ref{fig:stage} shows the effect of the proposed network-wise training, false positive and negative reduction and boundary smoothing. The top row shows that lung boundaries from trained model at stage 3 are much smoother than those from the model at stage 1. Also, the second and third rows support that false positives and false negatives can be supressed as stage goes. The performance plot in the right side in Fig.~\ref{fig:stage} shows the change of performances according to stages. It is observed that they are saturated at stage 3, so we report the performances from the model at stage 3.

In addition, we investigate the effect of data augmentation. For this, we adjust pixel values randomly through adjusting brightness and contrast so that the network is invariant to pixel value perturbations\footnote{Cropping, horizontal flipping and rotation were not effective. This is because lung segmentation network does not need to be invariant to such transformations.}. As shown in Table~\ref{table:comparison}, the trained model at stage 3 with data augmentation gives much better segmentation performances.

However, it should be noted that the performances on MC dataset are not as good as those on JSRT. Mean performances are slightly lower than the the hybrid approach in~\cite{Candemir2014TMI}, but standard deviations are much higher even if the model is trained with data augmentation. It means that our model traind using JSRT gives unstable segmentation results on some cases in MC as shown in Fig.~\ref{fig:examples}. This is due to the presence of a shift between distributions of training and test datasets, which needs to solve \textit{Domain Adaptation} problem. 

The samples of segmented lung boundaries are visualized in Fig.~\ref{fig:examples}. Left two columns show the best two results in terms of JSC, and right two columns show the worst two for JSRT and MC datasets.

%%\vspace{-3mm}

\section{Conclusion}

In this paper, we present an accurate lung segmentation model based on CNN with atrous convolutions. Furthermore, a novel multi-stage training strategy, network-wise training, to refine the segmentation results is also proposed. Computational experiments on benchmark dataset, JSRT, show that the proposed architecture and the network-wise training are very effective to obtain the accurate segmentation model for lung fields. We also evaluate the trained model on MC dataset, which raises the task for us to develop the model insensitive to domain shift.

{\small
\bibliographystyle{splncs03}
\bibliography{egbib}
}

\end{document}